\documentclass[letterpaper]{article} 
\usepackage{aaai23}  
\usepackage{times}  
\usepackage{helvet}  
\usepackage{courier}  
\usepackage[hyphens]{url}  
\usepackage{graphicx} 
\usepackage{natbib}  
\usepackage{caption} 
\usepackage{algorithm}
\usepackage{algorithmic}
\usepackage{adjustbox}
\usepackage{newfloat}
\usepackage{listings}
\DeclareCaptionStyle{ruled}{labelfont=normalfont,labelsep=colon,strut=off} 
\floatstyle{ruled}
\newfloat{listing}{tb}{lst}{}
\floatname{listing}{Listing}
\usepackage{xspace}
\usepackage{multirow}
\usepackage{enumitem}
\usepackage{amsmath}
\usepackage{pifont}
\usepackage{booktabs}
\newcommand{\etal}{\emph{et al.}\xspace}
\newcommand{\eg}{\emph{e.g.,}\xspace}
\newcommand{\ie}{\emph{i.e.,}\xspace}
\newcommand{\etc}{\emph{etc.}\xspace}
\newcommand{\baby}{AutoSTL\xspace}
\newcommand{\babylayer}{hidden layer\xspace}
\newcommand{\babylayers}{hidden layers\xspace}
\newcommand{\babylayercapital}{Hidden Layer\xspace}
\newcommand{\STDM}{spatio-temporal prediction\xspace}
\newcommand{\STDMcapital}{Spatio-temporal prediction\xspace}

\usepackage{color}

\newcommand{\zzj}[1]{{\color{black} #1}}

\title{AutoSTL: Automated Spatio-Temporal Multi-Task Learning}
\author {
    Zijian Zhang\textsuperscript{\rm 1,\rm 2,\rm 3,\rm 7},
    Xiangyu Zhao\textsuperscript{\rm 2,\rm 7}\thanks{Corresponding authors.},
    Hao Miao\textsuperscript{\rm 4},
    Chunxu Zhang\textsuperscript{\rm 1,\rm 3},
    Hongwei Zhao\textsuperscript{\rm 1,\rm 3},
    Junbo Zhang\textsuperscript{\rm 5,\rm 6}
}
\affiliations {
    \textsuperscript{\rm 1} College of Computer Science and Technology, Jilin University, China\\
    \textsuperscript{\rm 2} School of Data Science, City University of Hong Kong, Hong Kong\\
    \textsuperscript{\rm 3} Key Laboratory of Symbolic Computation and Knowledge Engineering of Ministry of Education, Jilin University, China\\
    \textsuperscript{\rm 4} Department of Computer Science, Aalborg University, Denmark\\
    \textsuperscript{\rm 5} JD Intelligent Cities Research, China\\
    \textsuperscript{\rm 6} JD iCity, JD Technology, China\\
    \textsuperscript{\rm 7} Hong Kong Institute for Data Science, City University of Hong Kong, Hong Kong\\
    \{zhangzj2114,cxzhang19\}@mails.jlu.edu.cn,
    xianzhao@cityu.edu.hk,
    haom@cs.aaudk,
    zhaohw@jlu.edu.cn,
    msjunbozhang@outlook.com
}

\usepackage{bibentry}

\begin{document}

\maketitle

\begin{abstract}
\zzj{\STDMcapital}
plays a critical role in smart city construction. 
Jointly modeling multiple spatio-temporal tasks can further promote an intelligent city life by integrating their inseparable relationship.
However, existing studies fail to address this joint learning problem well, which generally solve tasks individually 
or a fixed task combination.
The challenges lie in the tangled relation between different properties, the demand for supporting flexible combinations of tasks and the complex spatio-temporal dependency. 
To cope with the problems above, we propose an Automated Spatio-Temporal multi-task Learning (\baby) method to handle multiple spatio-temporal tasks jointly. Firstly, we propose a scalable architecture consisting of advanced spatio-temporal operations to exploit the complicated dependency. 
Shared modules and feature fusion mechanism are incorporated to further capture the intrinsic relationship between tasks.
Furthermore, our model automatically allocates the operations and fusion weight.
Extensive experiments on benchmark datasets verified that our model achieves state-of-the-art performance. As we can know, \baby is the first automated spatio-temporal multi-task learning method.
\end{abstract}

\section{Introduction}
With the conspicuous progress of data mining techniques, 
\zzj{\STDM}
has unprecedentedly facilitated today's society, such as traffic state modeling \cite{strn, wang2022inferring, zhou2020riskoracle, xu2016taxi}, 
urban crime prediction \cite{aaai22, zhao2017exploring, zhao2017modeling},
next point-of-interest recommendation \cite{guo2016cosolorec, cui2021st}, 
\etc
\zzj{\STDMcapital}
aims to model spatial and temporal patterns from historical spatio-temporal data, and predict the future states. 

\begin{figure}[!t]
	\includegraphics[width=1\linewidth]{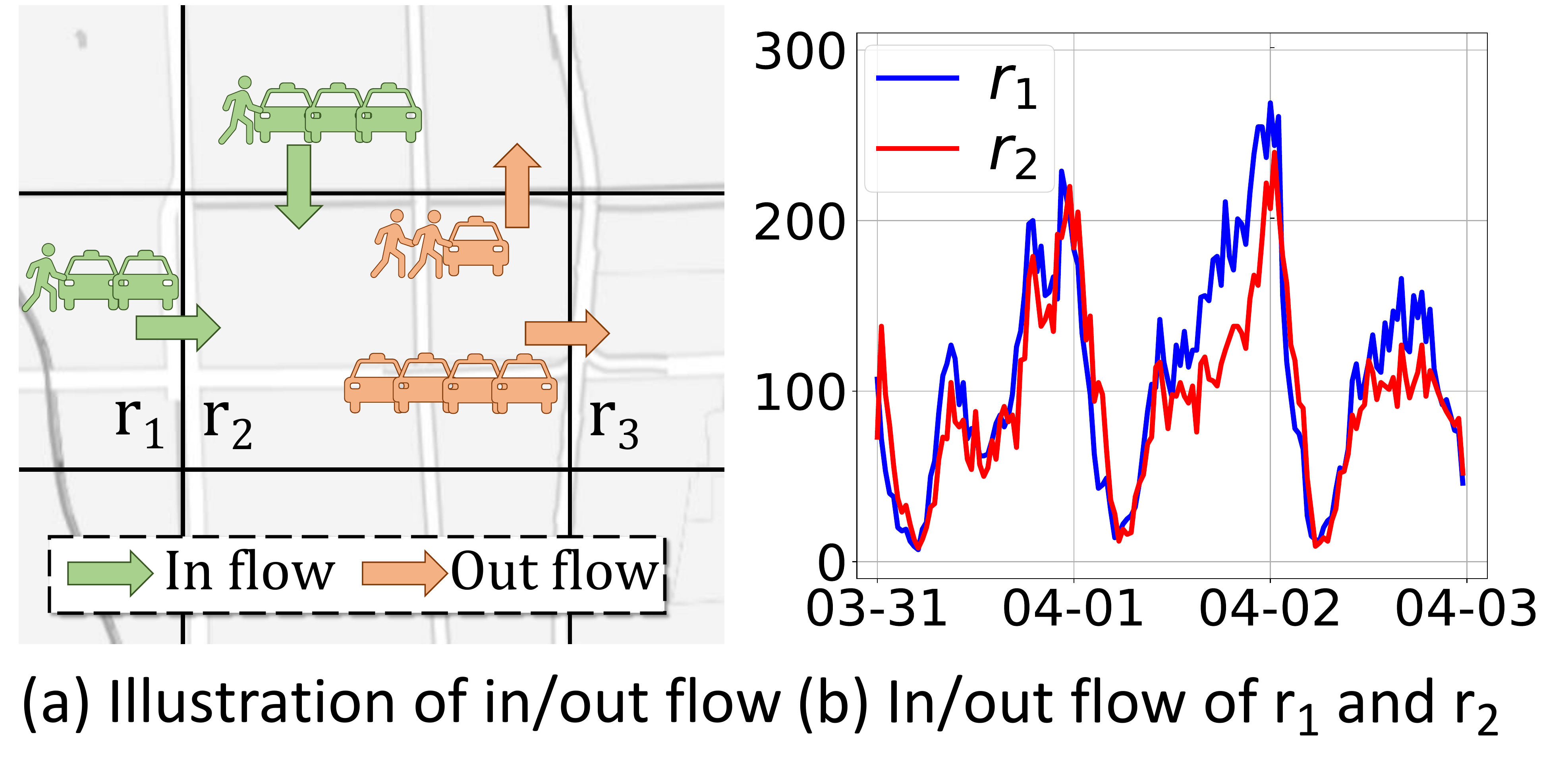}
	\caption{Illustration of close relationship between multiple spatio-temporal properties and regions.
	}
	\label{fig:intro}
\end{figure}

Generally, the 
\zzj{\STDM} tasks are defined and handled individually. Take traffic state modeling\zzj{, the focus of this paper,} as an example, the traffic state has been divided into multiple tasks, such as traffic flow prediction \cite{strn, ye2021coupled}, on-demand flow prediction \cite{feng2021multi}, traffic speed prediction \cite{dcrnn, gwnet}, \etc Most 
\zzj{\STDM} researches are engaged in pursuing higher model capacity on a single task.
Nevertheless, it is pivotal to propose an architecture that is capable of jointly handling multiple 
\zzj{\STDM} tasks.
The properties of different tasks evolve coherently,  and addressing the intrinsic information sharing between different tasks benefits each task \cite{caruana1997multitask}.
Firstly, different properties of a single region are highly related. As illustrated in Figure \ref{fig:intro}(a), traffic in flow and out flow of region $r_2$ are highly-correlated, which share consistent volume and fluctuation. 
On the other hand, different properties of different regions share similar characteristics. In Figure \ref{fig:intro}(b), in flow of $r_1$ and out flow of $r_2$ vary with similar periodicity and trend.
A similar phenomenon generally appears between multiple properties as well as nonadjacent regions. 
\zzj{Capturing multiple tasks together can benefit both efficiency and efficacy \cite{tang2020progressive, chen2022refined, zhao2022multi}.}

\begin{figure*}[!t]
	\centering
	\includegraphics[width=0.78\linewidth]{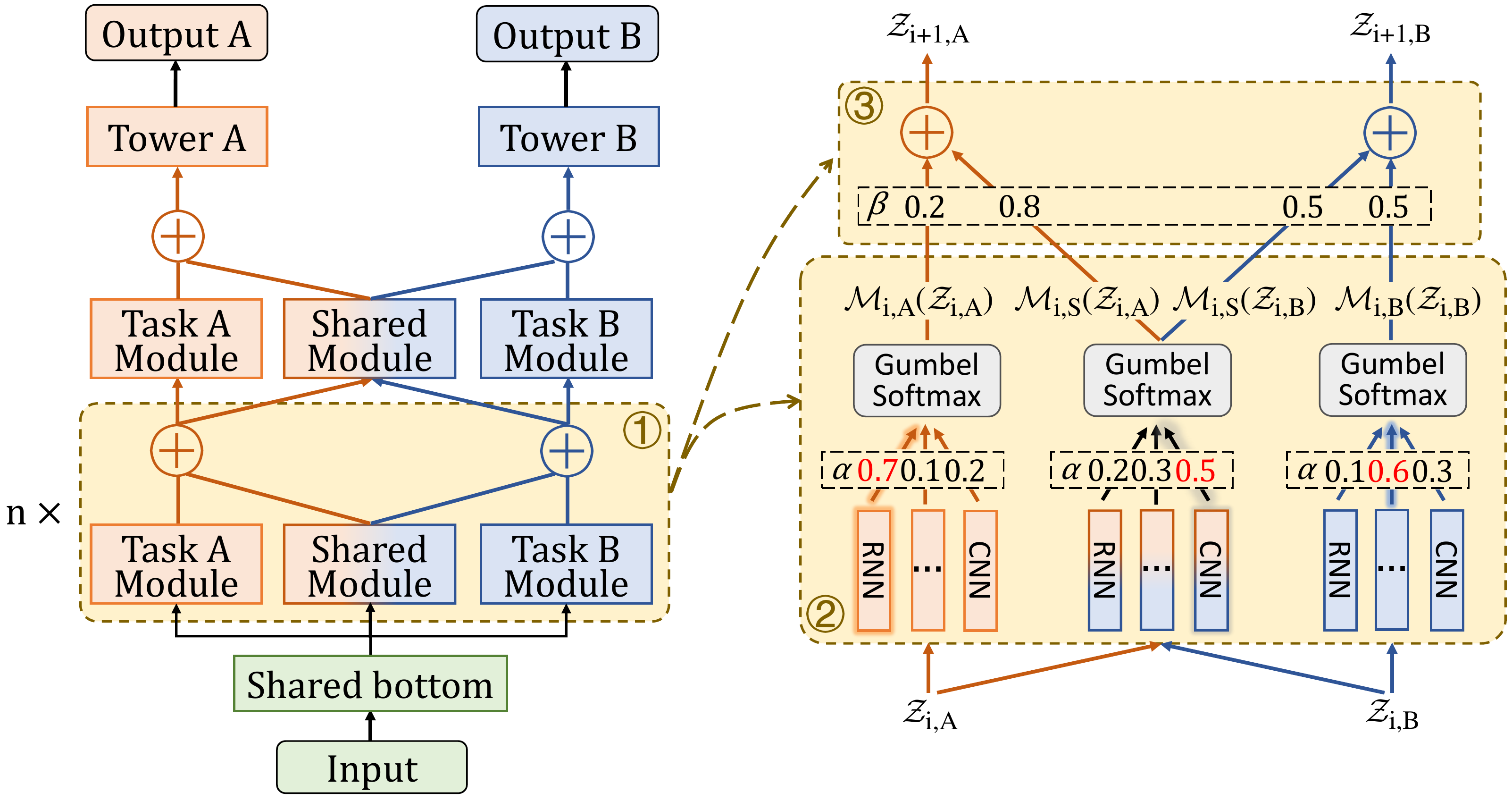}
	\caption{Model framework. The left part is the architecture of \baby. From bottom to top, the model consists of shared bottom, $n$ \babylayers and task-specific tower. \ding{172} represents the \babylayer. The right part is the illustration of \babylayer, which consists of task-specific and shared modules \ding{173} and module fusion mechanism \ding{174}. }  
	\label{fig:framework}
\end{figure*}
Researchers have been paying growing attention to Spatio-Temporal Multi-Task Learning (STMTL). The earliest attempt employs convolutional neural network to model the traffic flow and on-demand flow \cite{zhang2019flow}. LSTM-based method was also incorporated to predict traffic in flow and out flow simultaneously \cite{zhang2020taxi}. 
However, there are several limitations for existing STMTL methods. Firstly, relationship between different tasks has not been well-addressed. Current STMTL methods basically employ MLP fusion layer or concatenation to fuse different task features. Secondly, in order to adapt to multiple properties, these methods are specified with manually designed module and architecture such as stacked LSTM or CNN, which may not be optimal for the complex task correlation and introduce 
human error. Last, 
existing methods aim to fixed combination of tasks, which suffer from poor transferability and scalability. 
It demands reconstruction with tremendous expert efforts for other tasks or data. 


In this paper, we propose a self-adaptive architecture, named \baby, to solve multiple spatio-temporal tasks simultaneously. 
We mainly face several serious challenges. 
First, the spatio-temporal correlations of all grids are complex to be well-captured, especially modeling multiple spatio-temporal properties.
Besides, to benefit each specific task, it demands exploiting their entangled relationship and an appropriate feature fusion mechanism.
Furthermore, in order to support flexible multiple task combinations, the model is supposed to be scalable and easily extendable.
Finally, manually-designed model requires tremendous expert efforts and may trigger bias or mistakes, leading to a sub-optimal model. 
To fully address the problems above, 
we design a scalable architecture \baby, which can automatically allocate modules with a set of advanced spatio-temporal operations.
We introduce shared modules to specifically address the relationship between properties and an automated fusion mechanism to fuse multi-task features. The architecture consists of stacking \babylayers, where the modules inside and the fusion weight are fully self-adaptive. 
The main contributions could be summarized as follows:
\begin{itemize}[leftmargin=*]
    \item We propose a novel end-to-end framework, \baby, to solve multiple spatio-temporal tasks simultaneously. It can automatically choose the suitable saptio-temporal module and fusion weight with trivial cost. 
    \item A shared module modeling relationship between multiple tasks is incorporated, which can well-exploit the intrinsic dependency and benefit every single task.
    \item 
    Our proposed model \baby is the first model solving spatio-temporal multi-task learning automatically.
    {Extensive experiments with multiple multi-task settings verified its efficacy and generality.}
\end{itemize}

\section{Preliminaries}
\textbf{Problem Definition.}
\emph{ Spatio-Temporal Multi-Task Prediction.}
Let a graph $\boldsymbol{\mathcal{G}} = (\boldsymbol{\mathcal{V}}, \boldsymbol{\mathcal{E}})$ describe the traffic state data. $\boldsymbol{\mathcal{V}}$ is the node set representing regions of city, and $\boldsymbol{\mathcal{E}}$ is the edge set which depicts the connectivity between regions. 
At time step $t$, the multi-task feature matrix of $\boldsymbol{\mathcal{G}}$ is $\boldsymbol{X}_t=$
[$\boldsymbol{x}_{t,1},...,\boldsymbol{x}_{t,k},...,\boldsymbol{x}_{t,K}$]
$\in$
$\boldsymbol{R}^{N \times K}$,
where $N$ is the number of region, $K$ is the number of task, and 
$k = 1,...,K$ represents certain spatio-temporal task such as traffic speed, flow and trip duration of regions.


Spatio-temporal Multi-Task prediction aims to predict future multiple traffic states simultaneously given the historic states, by capturing the spatial and temporal variation pattern. The mapping function ${f_{\boldsymbol{W}}}$ parameterized by $\boldsymbol{W}$ is: 

\begin{equation}\label{Equ:Defination}
    \boldsymbol{x}_{1:T,k},\boldsymbol{\mathcal{G}} \stackrel{f_{\boldsymbol{W}}}{\longrightarrow} \boldsymbol{y}_k, \forall k=1,...,K
\end{equation}


\noindent where $\boldsymbol{x}_{1:T,k}$ $\in$ $\boldsymbol{R}^{T\times N}$ indicates the traffic state of historical $T$ time steps. $\boldsymbol{y}_k$ $\in$ $\boldsymbol{R}^{N}$ is the traffic state of task $k$ in the future time step. For convenient description, we omit the subscript and denote $\boldsymbol{X}_{1:T}$ as $\boldsymbol{X}$.


\section{Methodology}

\subsection{Architecture Overview}

We propose a scalable architecture to solve multiple spatio-temporal tasks in an automated way, named \baby. As visualized in Figure \ref{fig:framework}, \baby generally consists of shared bottom, \babylayers and task-specific tower. 

We employ scalable \babylayer to capture spatio-temporal dependency of multiple tasks simultaneously. Specifically, we incorporate task-specific module to model each property and shared module to capture the intrinsic relationship between tasks. 
For module design, we define a spatio-temporal operation set including recurrent neural network, convolution neural network, graph convolution network and Transformer. The concrete assignation is searched by Automated Machine Learning (AutoML).
For a further flexible fusion of multiple task properties, we propose self-adaptive fusion weights and optimize with AutoML.
After representation learning of \babylayers, the task-specific tower predict by feeding on the multiple task features. 

It is noteworthy that the module operations and fusion weights of \baby are automatically decided. The optimal allocation contributes to effectively capturing spatio-temporal patterns of specific task and intrinsic relationships between multiple spatio-temporal tasks.

\subsection{Framework}

\subsubsection{Shared Bottom}
We transform the input data with an MLP layer shared bottom. Specifically, we input the feature matrix of historic $T$ time steps $\boldsymbol{X}$, and let $\boldsymbol{\mathcal{Z}}_0$ represent the hidden representation learned by the shared bottom.
\begin{equation}\label{Equ:SharedBottom}
    \boldsymbol{\mathcal{Z}}_0 = \sigma(\boldsymbol{W}_s\boldsymbol{X}+\boldsymbol{b}_s)
\end{equation}
\noindent where $\boldsymbol{W}_s$ and $\boldsymbol{b}_s$ represent weight and bias of MLP layer respectively, and $\sigma$ is activation function.

\subsubsection{\babylayercapital}
We employ $n$ stacking \babylayers to capture the spatio-temporal dependency of each task and the intrinsic relationship between different tasks.
As shown in Figure \ref{fig:framework}, a \babylayer includes modules and fusion mechanism.
We incorporate task-specific module for each task to address spatio-temporal dependency, and shared module to address the intrinsic relationship between different tasks. 
For each module, we exploit a spatio-temporal operation set, and adaptively select one spatio-temporal operation from the operation set. Processed by the spatio-temporal operations, the features are flexibly fused by module fusion {mechanism}.

\paragraph{Spatio-Temporal Operation Set}
We maintain a spatio-temporal operation set including multiple spatio-temporal operations to support the automatic assignment of module operations. Specifically, we employ diffusion convolution \cite{dcrnn}, 1-D dilated causal convolution \cite{gwnet}, long-short term memory (LSTM), informer \cite{informer} and spatial-informer \cite{wu2021autocts} as spatio-temporal data mining operations, abbreviated as GCN, CNN, RNN, TX, and {TX\_S, respectively}.

Diffusion convolution has an edge on capturing spatial dependency in data. In particular, it characterizes graph convolution based on $K$-step random walking on the graph. The diffusion convolution on the hidden representation is:
\begin{equation}
\label{Equ:dcrnn}
\begin{aligned}
    \!\mathcal{O}_{GCN}\!(\boldsymbol{\mathcal{Z}}_i) \!= \!\sum_{k=0}^{K}\!\left(\boldsymbol{D}_{O}^{-1} \!\boldsymbol{A}\right)^{k} \!\boldsymbol{\mathcal{Z}}_i \!\boldsymbol{W}_{1,k}\!+\!\left(\!\boldsymbol{D}_{I}^{-1} \!\boldsymbol{A}^{\top}\!\right)^{k} \!\boldsymbol{\mathcal{Z}}_i \!\boldsymbol{W}_{2,k}
\end{aligned}
\end{equation}
\noindent where $\boldsymbol{A}$ is the adjacent matrix of $\boldsymbol{\mathcal{G}}$, $\boldsymbol{D}_{O}^{-1}$ and $\boldsymbol{D}_{I}^{-1}$ are out-degree and in-degree diagonal matrices. $\boldsymbol{W}_{1,k}$ and $\boldsymbol{W}_{2,k}$ are trainable {filter parameters}.


1-D dilated causal convolution is an effective operation for temporal information. Through padding zero to input data, it reserves causal temporal dependency and is able to predict next state based on past states \cite{gwnet}.
\begin{equation}
\label{Equ:dcc}
\mathcal{O}_{CNN}(\boldsymbol{\mathcal{Z}}_i)=\left(\boldsymbol{\mathcal{Z}}_i \boldsymbol{W}_{3}\right) \odot \sigma\left(\boldsymbol{\mathcal{Z}}_{i} \boldsymbol{W}_{4}\right)
\end{equation}

LSTM is a classical technique to learn temporal information. It enjoys a naturally-progressive architecture to model the sequential dependency and predict the future state. It is generally-applied in spatio-temporal data mining methods \cite{zhang2016dnn, strn}.
\begin{equation}
\label{Equ:lstm}
\mathcal{O}_{RNN}(\boldsymbol{\mathcal{Z}}_i) =LSTM(\boldsymbol{\mathcal{Z}}_i)
\end{equation}

Transformer has proven its efficacy in recent 
\zzj{\STDM} advances \cite{guo2019attention, li2019enhancing}, we employ its efficient variant \textit{informer} \cite{informer}, and the enhanced version \textit{spatial-informer} considering spatial dependency \cite{wu2021autocts} as operations, \ie TX and TX\_S. TX could be formulated as follows:
\begin{equation}
    \label{Equ:transformer}
\mathcal{O}_{TX}\!=\!{softmax}\bigg(\!
\frac{\psi \big(\boldsymbol{\mathcal{Z}}^{(i)} \boldsymbol{W}_{Q}\big)\big(\boldsymbol{\mathcal{Z}}^{(i)} \boldsymbol{W}_{K}\big)^{\top}}
{\sqrt{d}}
\!\bigg)
\big(\boldsymbol{\mathcal{Z}}^{(i)} \boldsymbol{W}_{V}\big)
\end{equation}

\noindent where $\psi$ is the sampling function in \textit{informer}, $d$ is feature dimension, and $\boldsymbol{W}_Q$, $\boldsymbol{W}_K$, $\boldsymbol{W}_V$ are trainable weight matrices. Similarly,  \textit{spatial-informer} conducts attention mechanism on spatial relationship, \ie $\mathcal{O}_{TX\_S}$ feeds on the spatial transposition of $\boldsymbol{\mathcal{Z}}_i$. 

Based on the spatio-temporal operations, we attain the operation set
 $\boldsymbol{\mathcal{O}}$ = [$\mathcal{O}_{GCN}$, $\mathcal{O}_{RNN}$, $\mathcal{O}_{CNN}$, $\mathcal{O}_{TX}$, $\mathcal{O}_{TX\_S}$].
\zzj{As in bottom right of Figure \ref{fig:framework}, operations in each module are from $\boldsymbol{\mathcal{O}}$.}
It is worth noting that the spatio-temporal operation set is extensible. We select operations due to their advanced capacity of modeling spatial and temporal dependency as well as efficiency. The operation set could be easily modified or extended for future potential enhancement.

\begin{figure*}[!ht]
\centering
	\includegraphics[width=0.9\linewidth]{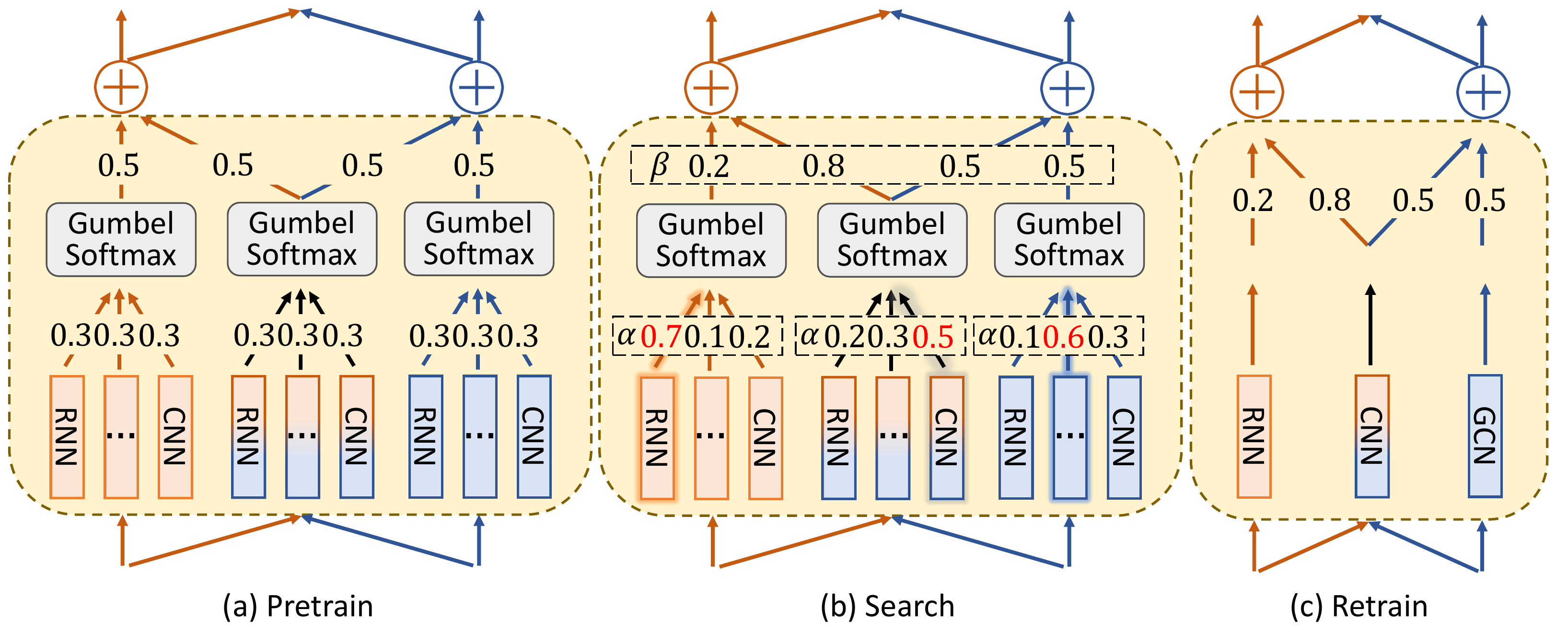}
	\caption{Three phases of weights searching. (a). Pretrain architecture, weights are fixed as initialization. (b). Search optimal weights by AutoML. (c). Identify operation in modules and achieve the optimal architecture.}
	\label{fig:automl}
\end{figure*}

\paragraph{Task-specific and Shared Module}
As visualized in Figure \ref{fig:framework}, we present two kinds of modules in \babylayer, \ie task-specific module and shared module. Task-specific module aims to capture information benefiting a certain task, and shared module is supposed to learn the entangled correlation between different tasks. Note that in each \babylayer, we can assign multiple task-specific modules and shared modules. For clear description, we assign one task-specific module for each task as well as one shared module in each \babylayer, as shown in Figure \ref{fig:framework}. 
We denote the module of layer $i$ as $\mathcal{M}_{i,\lambda}$, where $\lambda$$\in$ \{A,B,S\}. $\lambda=S$ stands for shared module, and $\lambda=A$ and $B$ represent task-specific module of task A and B, respectively. For instance, $\mathcal{M}_{2,A}$ is the specific module of task A in the $2$-nd \babylayer. 

To maintain a trade-off between efficacy and efficiency, we search one specific operation for each module, rather than utilize all operations. 
There emerges a challenge that the hard selection of operations can not be optimized with gradient due to the non-differentiable selection. Conventionally, a soft weight $\boldsymbol{\alpha}$ is added to depict the significance of candidate operations, and a soft fusion is hired to approach the selection. But it can hardly avoid reaching the sub-optimal result.
To overcome this, we incorporate gumbel-softmax to approximate the hard selection of {operations}. 

To identify operation in module, we assign weight for each module $\boldsymbol{\alpha} = $ [$\alpha^{GCN}$, $\alpha^{RNN}$, $\alpha^{CNN}$, $\alpha^{TX}$, $\alpha^{TX\_S}$] to weight all candidate operations.
Then the operations of module could be identified through hard sampling by gumbel-max technique \cite{gumbel1954statistical}. 
For instance, the module of the task A $\mathcal{M}_{i,A}(\boldsymbol{\mathcal{Z}}_{i,A})$ could be reached by:
\begin{equation}
\label{Equ:taskspecific_1}
\mathcal{M}_{i,A}(\boldsymbol{\mathcal{Z}}_{i,A})\! = \!one\_{hot}\left(\underset{j \in[1, |\boldsymbol{\mathcal{O}}|]}{argmax}\left[\log \alpha_{j}+g_{j}\right]\right) \odot \boldsymbol{\mathcal{O}}(\boldsymbol{\mathcal{Z}}_i)
\end{equation}

\noindent where $g_j=-log(-log(\varepsilon _j))$ and $\varepsilon_j $ belongs to uniform distribution. $\odot$ is the dot product.

Due to the non-differentiable argmax operation, we introduce Gumbel-softmax \cite{jang2016categorical, zhao2021autoloss}. In particular, it simulates hard selection based on reparameterization on categorical distribution:
\begin{equation}
\label{Equ:onehot}
p_{j}=\frac{\exp \left(\left(\log \left(\alpha_{j}\right)+g_{j}\right) / \tau\right)}{\sum_{k=1}^{n} \exp \left(\left(\log \left(\alpha_{k}\right)+g_{k}\right) / \tau\right)}
\end{equation}
\noindent where $p_j$ is the probability of selecting the operation $j$, and $\tau$ controls the approach to hard selection. When $\tau$ approaches to zero, the Gumbel-softmax outputs {one-hot vector}.

\paragraph{Module Fusion Mechanism}
In order to model both task-specific and cross-tasks information, we propose a flexible module fusion mechanism. 
Take task A as an example, 
we fuse the output of its task-specific module $\mathcal{M}_{i,A}$ and shared module $\mathcal{M}_{i,S}$, and obtain the task-specific feature $\boldsymbol{\mathcal{Z}}_{i+1,A}$. Then, we output the task-specific feature to the task-specific and shared-modules in the next \babylayer. 
To dynamically weight each module's contribution to the output, we set the fusion weight $\boldsymbol{\beta}=[\boldsymbol{\beta}_A, \boldsymbol{\beta}_B]$ and search through gradient. So the task-specific feature of task A and B learned by the $(i+1)$-th layer $\boldsymbol{\mathcal{Z}}_{i+1,A}$ and $\boldsymbol{\mathcal{Z}}_{i+1,B}$ are as follows, which is illustrated in Figure \ref{fig:framework}: 
\begin{equation}
\label{Equ:fusion_specific1}
\boldsymbol{\mathcal{Z}}_{i+1,A}=softmax(\boldsymbol{\beta}_A) \odot [\mathcal{M}_{i,A}(\boldsymbol{\mathcal{Z}}_{i,A}),\mathcal{M}_{i,S}(\boldsymbol{\mathcal{Z}}_{i,A})]
\end{equation}
\begin{equation}
\label{Equ:fusion_specific2}
\boldsymbol{\mathcal{Z}}_{i+1,B}=softmax(\boldsymbol{\beta}_B) \odot [\mathcal{M}_{i,B}(\boldsymbol{\mathcal{Z}}_{i,B}),\mathcal{M}_{i,S}(\boldsymbol{\mathcal{Z}}_{i,B})]
\end{equation}

Notably, the input of both tasks of the first \babylayer is $\boldsymbol{\mathcal{Z}}_0$.
As shown in the left bottom of Figure \ref{fig:framework}, the output of shared bottom are duplicated and input to the first \babylayer, and the intermediate \babylayers feed on task specific features.
The task-specific data output by the last \babylayer goes through task-specific tower and achieves the prediction {of each task}.


\subsubsection{Task-specific Prediction Tower}
After representation learning of $n$ \babylayers, we utilize task-specific tower to predict the future state. We hire MLP layers as tower, which enjoys low computational complexity.
\begin{equation}
\label{Equ:tower1}
\boldsymbol{Y}_{A} = \boldsymbol{W}_{oA}(\boldsymbol{\mathcal{Z}}_{n,A})+\boldsymbol{b}_{oA}
\end{equation}
\begin{equation}
\label{Equ:tower2}
\boldsymbol{Y}_{B} = \boldsymbol{W}_{oB}(\boldsymbol{\mathcal{Z}}_{n,B})+\boldsymbol{b}_{oB}
\end{equation}

\noindent where $\boldsymbol{W}_{oA}$ and $\boldsymbol{b}_{oA}$ are the weight matrix and bias of the tower of task A, so do $\boldsymbol{W}_{oB}$ and $\boldsymbol{b}_{oB}$ to task B.

\begin{table*}[!t]
\centering
\small
\renewcommand{\arraystretch}{1.0}
\begin{tabular}{c||cc||cc||cc||cc||cc||cc} 
\hline
\multicolumn{1}{c||}{\textbf{Dataset}} & \multicolumn{8}{c||}{\textbf{NYC Taxi}} & \multicolumn{4}{c}{\textbf{PEMSD4}}\\ \hline
\multicolumn{1}{c||}{Metrics} & \multicolumn{2}{c||}{RMSE} & \multicolumn{2}{c||}{MAE} & \multicolumn{2}{c||}{RMSE} & \multicolumn{2}{c||}{MAE} & \multicolumn{2}{c||}{RMSE} & \multicolumn{2}{c}{MAE} \\ 

\multicolumn{1}{c||}{Methods} & In & Out & In & Out & OD & Duration & OD & Duration & Flow & Speed & Flow & Speed\\
\hline
ARIMA & 23.63 & 25.36 & 13.81 & 13.87 & 1.04 & 3.21 & 0.84 & 2.01 & 146.01 & 8.37 & 114.51 & 4.40\\
DCRNN & 7.75 & {7.21} & 4.73 & 4.58 & 0.63 & 3.09 & 0.25 & 1.52 & 27.67 & 2.06 & \underline{17.42} & \underline{1.11}\\
GWNet & 8.89 & 7.27 & 5.37 & 4.60 & \underline{0.60} & {2.90} & \underline{0.23} & {1.24} & 30.61 & 2.27 & 19.62 & 1.28 \\
CCRNN & 8.04 & 7.35& 4.87 & 4.52  & 0.61 & 2.97 & \underline{0.23} & 1.40 & 28.13 & 2.07 & 17.81 & 1.14\\
\hline
PLE & 8.77 & 8.46 & 5.29 & 5.19 & 0.66 & 3.05 & 0.27 & 1.44 & 30.14 & 2.25 & 19.24 & 1.29\\
PLE-LSTM & 8.66 & 8.31 & 5.19 & 5.09 & 0.65 & 3.04 & 0.27 & 1.38 & 29.60	& 2.11 & 18.66 & 1.20 \\
DCRNN-MTL & 7.85 & 7.52 & 4.84 & 4.75 & {0.62} & {2.92} & 0.24 & 1.28 & \underline{27.65} & 2.34 & {17.44} & 1.46\\
CCRNN-MTL & 7.68 & 7.50 & \underline{4.62} & 4.53 & 0.63 & \textbf{2.86} & 0.24 & \underline{1.22} & 27.99 & 2.56 & 17.59 & 1.52\\
GTS\dag & 7.86 & 7.40 & 4.90 & 4.65 & - & - & - & - & 28.05 & 2.09 & 17.87 & 1.15\\
MTGNN & \underline{7.64} & \underline{7.17} & 4.67 & \underline{4.48} & \textbf{0.59} & \underline{2.89} & {0.24} & {1.24} & 28.03 & \underline{2.05} & 17.92 & {1.12}\\
\hline

 \baby & \textbf{7.57*} & \textbf{7.16} & \textbf{4.53*} & \textbf{4.41*} & \textbf{0.59} & \underline{2.89} & \textbf{0.21*} & \textbf{1.12*}  & \textbf{27.36*} & \textbf{2.01*} & \textbf{17.40*} & \textbf{1.08*}\\
\hline

\end{tabular}
\dag GTS fails to run on traffic OD flow and trip duration due to its tremendous allocation of GPU space.
\caption{Overall experiment results. Best performances are bold, next best performances are underlined.
``\textbf{{\Large *}}'' indicates the statistically significant improvements (i.e., two-sided t-test with $p<0.05$) over the best baseline.}
\label{table:inoutflow}
\end{table*}

\subsection{Optimization by AutoML}
Traditional \zzj{\STDM} works endeavor to manually design specified architecture, which highly depends on expert's experience and suffers from poor generality. 
AutoML has demonstrated its efficiency and efficacy on searching model operation and architecture with advanced model capacity \cite{pan2021autostg, wu2021autocts,zhao2021autodim}.
We utilize AutoML to maintain self-adaptive model architecture as well as module operations.

The optimization procedure consists of 3 phases, \ie pretrain, search and retrain, as shown in Figure \ref{fig:automl}. First, we initialize and fix all searching weights $\boldsymbol{\alpha}$ and $\boldsymbol{\beta}$ as average value, and pretrain the model for several epochs. Then, we update $\boldsymbol{\alpha}$ and $\boldsymbol{\beta}$ with a mini-batch of validation data along with the model training, \ie Bi-level Optimization. Finally, after the search phase, the optimal operations in module and weight of fusion mechanism are already identified. We retrain the optimal architecture with a full-training.

\subsubsection{Bi-level Optimization}
In \baby, 
the concrete operations of modules in \babylayer and fusion weight are identified adaptively based on model training. AutoML will offer the model with optimal performance within the search space. Let $\boldsymbol{W}$ represent the neural network parameters of the \baby, $\boldsymbol{\alpha}$ and $\boldsymbol{\beta}$ stand for the operations in modules and fusion weights, respectively. The whole algorithm could be optimized with a bi-level optimization. 
Notably, the update of architecture parameters $\boldsymbol{\alpha}$ and $\boldsymbol{\beta}$ are based on a mini-batch of validation data, which avoids overfitting problem 
with an acceptable {computational cost}:
\begin{equation}
\label{Equ:optimization}
\begin{aligned}
&\min _{\boldsymbol{\alpha},\boldsymbol{\beta}} \mathcal{L}_{va l}\left(\boldsymbol{W}^{*}(\boldsymbol{\alpha},\boldsymbol{\beta}), \boldsymbol{\alpha},\boldsymbol{\beta}\right) \\
&\text { s.t. } \boldsymbol{W}^{*}(\boldsymbol{\alpha},\boldsymbol{\beta})=\arg \min _{\boldsymbol{W}} \mathcal{L}_{train}\left(\boldsymbol{W}, \boldsymbol{\alpha}^{*},\boldsymbol{\beta}^{*}\right)
\end{aligned}
\end{equation}
To alleviate the computational cost of the optimization of $\boldsymbol{W}^{*}(\boldsymbol{\alpha},\boldsymbol{\beta})$, we propose to approximate the inner optimization function with one step of gradient descent:
\begin{equation}
\label{Equ:optimization_inner}
\boldsymbol{W}^{*}(\boldsymbol{\alpha},\boldsymbol{\beta})\approx \boldsymbol{W}-\eta\partial_{\boldsymbol{W}} \mathcal{L}_{train}(\boldsymbol{W}, \boldsymbol{\alpha},\boldsymbol{\beta})
\end{equation}
\noindent where $\eta$ is learning rate.
Through iteratively minimizing training loss $\mathcal{L}_{train}$ and validation loss $\mathcal{L}_{val}$, we can achieve a model with the optimal performance. 

\section{Experiment}
In this section, we present the experiment result as well as analysis to verify the efficacy of our proposed \baby. 

\subsection{Datasets}
We evaluate \baby 
on two commonly used real-world benchmark datasets of \zzj{\STDM}, \ie \textbf{NYC Taxi}\footnote{https://www1.nyc.gov/site/tlc/about/tlc-trip-record-data.page} and \textbf{PEMSD4}\footnote{http://pems.dot.ca.gov/}.
We collect data from April to June in 2016 for NYC Taxi with 35 million trajectories, and the data of January and February in 2018 for PEMSD4. 


\subsection{Data Preprocessing}
To thoroughly prove the capability of \baby on STMTL, we propose multiple experimental scenarios with different datasets. In a nutshell, we execute two groups of multi-task on NYC Taxi, \ie traffic in flow and out flow, as well as traffic on-demand flow and trip duration. Besides, we propose one multi-task setting on PEMSD4, \ie traffic flow and speed of sensors on road. 

\subsection{Baselines}
We compare with two lines of representative \zzj{\STDM} methods, which can be grouped as methods for single task and multiple tasks. \textit{Methods for single task}:
\textbf{ARIMA} \cite{box2015time}, \textbf{DCRNN } \cite{dcrnn}, \textbf{GWNet } \cite{gwnet}, \textbf{CCRNN } \cite{ye2021coupled}. \textit{Methods for multiple tasks}: \textbf{PLE } \cite{tang2020progressive}, \textbf{PLE-LSTM }, \textbf{DCRNN-MTL}, \textbf{CCRNN-MTL}, \textbf{MTGNN } \cite{wu2020connecting}, \textbf{GTS } \cite{shang2021discrete}. 

\subsection{Experimental Setups}
To facilitate the reproducibility, we detail the experimental setting including training environment and implementation details.
We predict the traffic attribute of the future 1 time interval based on the historical 12 time steps, \ie $|T|=12$. We select root mean squared error (RMSE) and mean absolute error (MAE) as evaluation metrics.
All experimental results are the average value of 5 individual runs.

In terms of model structure, we assign 1 task-specific module and 1 shared module in each \babylayer, \eg 3 modules in one \babylayer for two-tasks learning.
We stack 3 \babylayers in total. 


\subsection{Overall Performance}
Table \ref{table:inoutflow} presents the overall experiment results. From the result we can exactly reach conclusions below:
In the group of the method for single-task, $\emph{(1)}$ ARIMA gets worse results than deep learning-based models in all tasks. The performance gap is because of the advanced capacity of neural networks. 
$\emph{(2)}$ DCRNN and GWNet take the leading place among the GCN-based methods, and DCRNN performs steadily on different datasets. The possible reason is that the adaptive adjacent matrix equipped with GWNet and CCRNN makes the performance unstable. 
$\emph{(3)}$ \baby outperforms single task baselines consistently, including the state-of-the-art spatio-temporal prediction methods, \ie DCRNN, GWNet, and CCRNN. It verifies \baby the advanced ability of capturing spatio-temporal dependency between different properties, which benefits each single task and attains the best performance.

Among the multi-task methods, $\emph{(1)}$ PLE-LSTM performs better than PLE, which shows the capability of LSTM for modeling temporal dependency beyond MLP.
$\emph{(2)}$ DCRNN-MTL and CCRNN-MTL did not achieve better results than their single-task setting. This is a seesaw phenomenon that model improves the performance of one task but hurts the other one, which commonly emerges in multi-task learning and is ascribed to the incapability of modeling multiple properties. 
$\emph{(3)}$ \baby achieves superior results than GTS and MTGNN, the latter two are designed specifically for modeling multi-variate time series.
It verifies the efficacy of the self-adaptive architecture and operations in \baby. 
$\emph{(4)}$ \baby attains consistently advanced performance while baseline models achieve fluctuant results on different tasks and datasets. 
The flexible spatio-temporal operation and the automatic allocation by AutoML contribute to the outstanding adaptivity to different tasks {and datasets}.

\begin{table}[t]
\centering

\small
\renewcommand{\arraystretch}{1.0}
\begin{tabular}{c|cc|cc} 
\hline

\multirow{2}{*}{Methods} & \multicolumn{2}{c|}{RMSE} & \multicolumn{2}{c}{MAE} \\
& In & Out & In & Out\\
\hline
w/o $\boldsymbol{\alpha}$ & 7.77 & 7.20 & 4.65 & 4.44 \\
\hline
w/o $\boldsymbol{\beta}$ & 7.94 & 7.47 & 4.75 & 4.59\\
\hline
w/o s-m & 9.13 & 7.22 & 5.69 & 4.51\\
\hline
\baby & \textbf{7.57} & \textbf{7.16} & \textbf{4.53} & \textbf{4.41} \\
\hline
\end{tabular}
\caption{Components analysis of \baby on NYC Taxi.} 
\label{table:ablation}
\end{table}

\subsection{Ablation Study}
In this subsection, we present several variants of our proposed method and make a detailed comparison with them to verify the effective components in \baby.

\begin{itemize}[leftmargin=*]
    \item \textbf{w/o $\boldsymbol{\alpha}$}: Fix operations in \babylayers with optimal ones.
\end{itemize}

\begin{itemize}[leftmargin=*]
    \item \textbf{w/o $\boldsymbol{\beta}$}: Fix fusion weights with average value.
\end{itemize}

\begin{itemize}[leftmargin=*]
    \item \textbf{w/o s-m}: Remove shared module in \babylayer.
\end{itemize}

Table \ref{table:ablation} presents the results of \baby and the variants. 
From the result, we can safely draw conclusions as follows:
$\emph{(1)}$ Based on the experiment result on NYC Taxi, we fix the operations of 3 \babylayers with GCN, GCN and RNN from bottom to top, which is a proper approximation to the manual-design model. 
According to the results,
we find that the specific operation allocation inside each \babylayer still affects the performance, and the automated architecture, \ie \baby, performs better.
$\emph{(2)}$ Through fixing the fusion weights $\boldsymbol{\beta}$ and considering the equal contribution of each module in the \babylayer, we test the function of the self-adaptive fusion mechanism. The results demonstrate the prominence of properly weighting different spatio-task features.
$\emph{(3)}$ By removing the shared module in each \babylayer, 
we test its contribution to multi-task learning.

\begin{figure}[t]
     \centering
    	\includegraphics[width=1\linewidth]{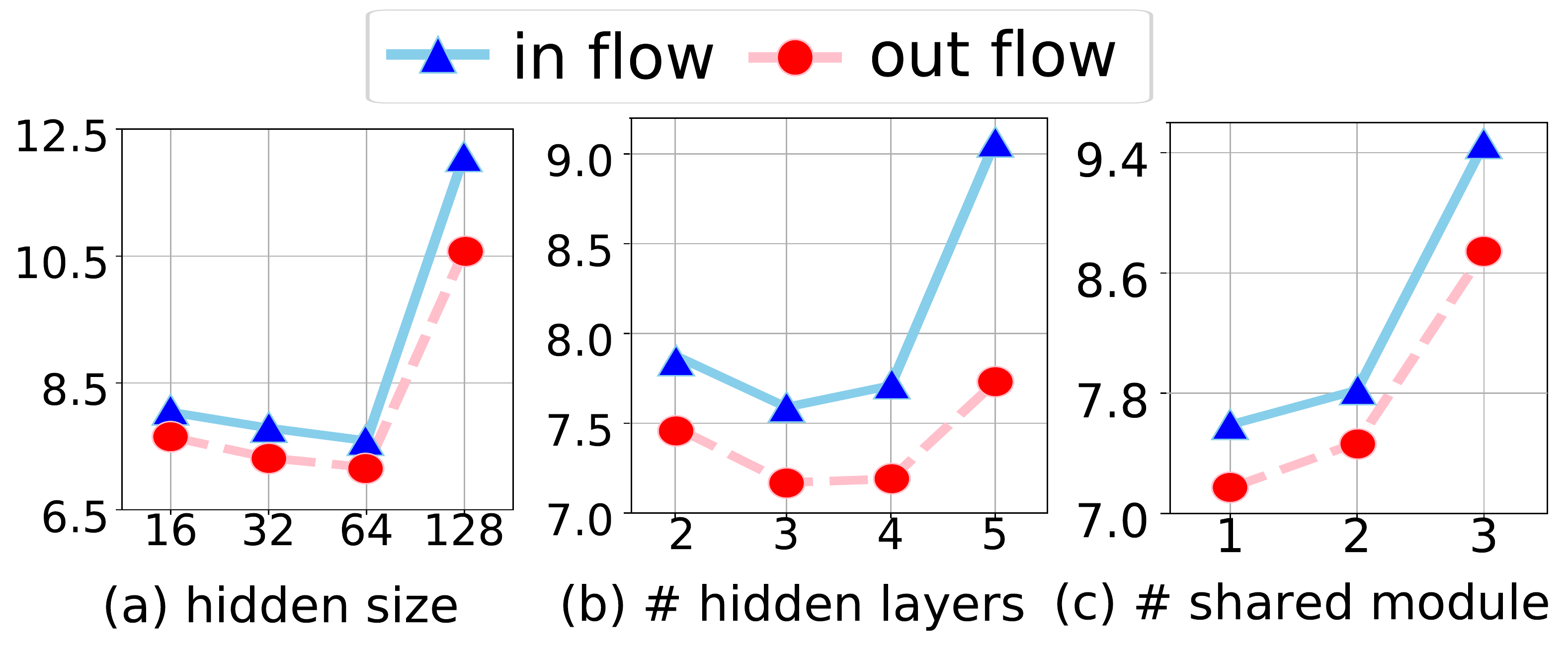}
    \caption{Influence of hyper-parameters on NYC Taxi dataset. We present the RMSE of traffic in and out flow.
    }
    \label{fig:hyper}
\end{figure}
The distinct performance gap below \baby proves the advanced ability to address the dependency of the shared module. 
Without
effectively modeling the relationship between multiple tasks,
w/o s-m converges with optimizing single task, \ie traffic out flow, whereas performs considerably less well in traffic in flow prediction.

\subsection{Hyper-parameter Analysis}
We demonstrate how 
hyper-parameters influence the performance of \baby.
We verify the key hyper-parameters, \ie the hidden size of embedding layer, the number of hidden layers and the number of shared module in hidden layer. 

Figure \ref{fig:hyper} presents the RMSE performance of traffic in and out flow on NYC Taxi dataset.
In Figure \ref{fig:hyper}(a), we test hidden size in \{16, 32, 64, 128\}. 
An improvement of hidden size in a relatively-low range can benefit performance, but when it is too large, \ie 128, the model collapses dramatically, which is possibly caused by overfitting problem.
From Figure \ref{fig:hyper}(b), \baby with 3 \babylayers achieves best performance,  fewer \babylayer impairs model capacity, while large ones may trigger overfitting.
As shown in Figure \ref{fig:hyper}(c), 
we can conclude that only 1 shared module in each \babylayer is enough for multiple properties modeling.

\subsection{Efficiency Comparison}
We compare parameter volume, training time, and inference time on NYC Taxi in Table \ref{table:parameter}.
For \baby, it contains 3 phases, \ie pretrain, search and retrain. 
In pretrain and search phases, the model has 142K parameters, and the final searched model has 312K for retrain phase.
Following Wu \etal \cite{wu2021autocts},
we set the embedding size of the first two phases a quarter of that of the retrain phase, so the pretrain and search phases have fewer parameters and take trivial training costs, \ie about one-seventh training cost of model retraining.
From the results, 
we can observe that \baby achieves state-of-the-art prediction effectiveness with competitive space and time consumption. 
MTGNN takes less training time than \baby, but it demands twice space allocation due to its MLP components.

\subsection{Visualization}
\begin{figure}[t]
	\centering
	\includegraphics[width=0.8\linewidth]{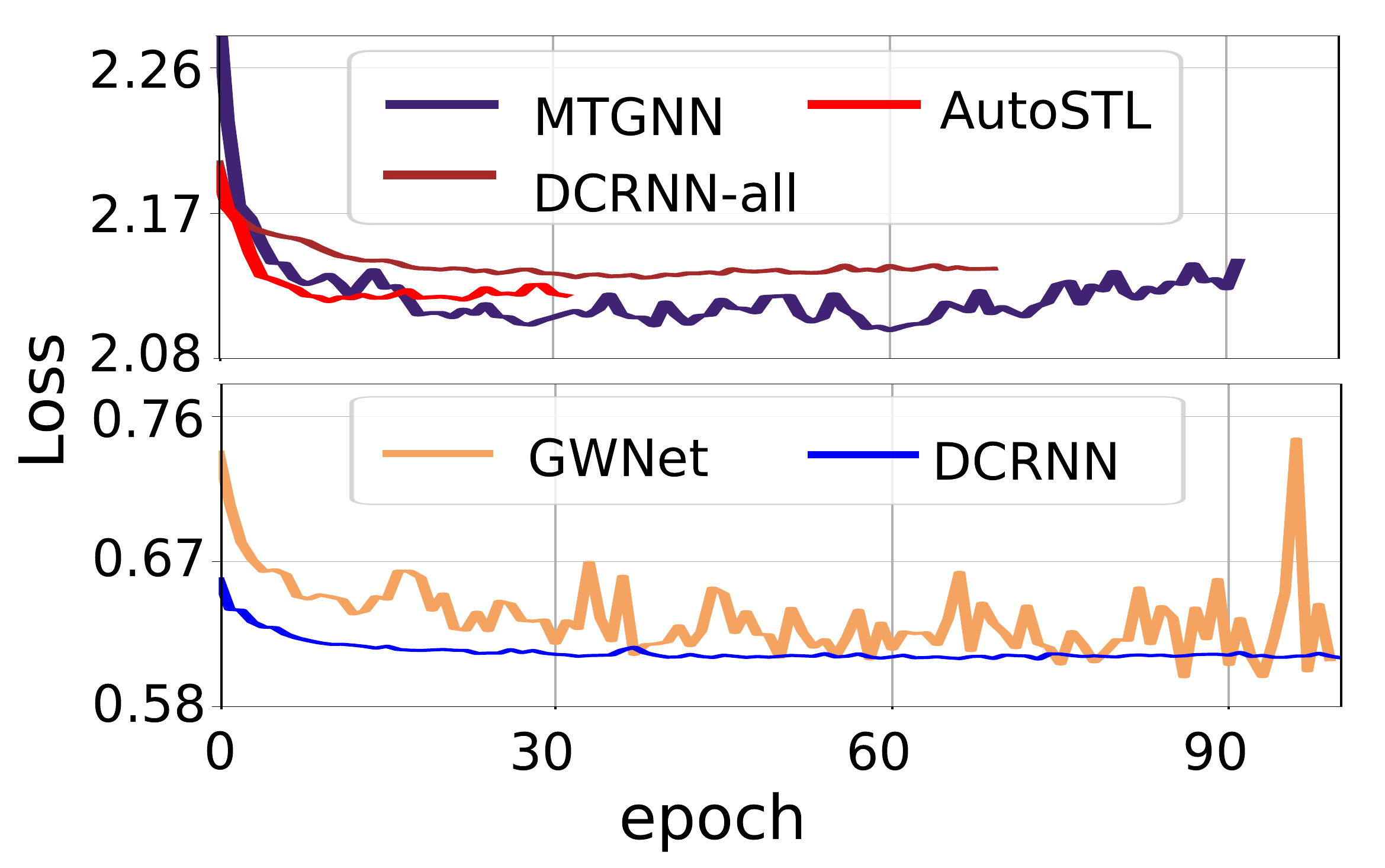}
	\caption{Loss curves comparison on NYC Taxi.
	}
	\label{fig:visual}
\end{figure}

We show the efficacy of \baby from multiple views. 
Figure \ref{fig:visual} illustrates the validation loss of traffic in flow on NYC Taxi with respect to training epoch. 
We can observe that \baby converge with least epochs, \ie 32, while all baselines take at least 70 epochs to converge.
\baby hires advanced spatio-temporal operations into a compact architecture, 
which leads to more accurate gradient descent, and fosters quicker convergence with fewer training epochs.

\begin{table}[t]
\centering
\small
\renewcommand{\arraystretch}{1.0}
\begin{tabular}{c||c|c|c|c|c} 
\hline
\multirow{2}{*}{Methods} & \multicolumn{2}{c|}{MAE} & {Para} & Training & Inference\\
& In & Out &(K) & time(s) & time(ms)\\
\hline
DCRNN & 4.73 & 4.58 & 127 & 11,556 & 1280 \\
GWNet & 5.37 & 4.60 & 272 & 2,056 & 40\\
CCRNN & 4.87 & \underline{4.52} & 139 & 2,218 & 53 \\
\hline
DCRNN-MTL & 4.84 & 4.75 & 127 & 5,783 & 1280 \\
CCRNN-MTL  & \underline{4.62} & 4.53 & 139 & {1,236} & 52\\
MTGNN & 4.67 & 4.48 & 612 & 1683 & 78\\
\hline
\baby & \textbf{4.53} & \textbf{4.41} & 312 & {1,958} & 100 \\
\hline
\end{tabular}
\caption{Space and time efficiency comparison.}
\label{table:parameter}
\end{table}






\section{Related Work}
\subsection{Traditional Spatio-Temporal Prediction}

Varieties of deep learning techniques have been applied to  
\zzj{\STDM}. The capability of deep learning techniques can be roughly divided into two categories, \ie temporal pattern capture and spatial pattern capture.  
\emph{For temporal pattern capture}, since Ma \etal \cite{ma2015long} and Tian \etal \cite{tian2015predicting} first applied LSTM to 
\zzj{\STDM}, there emerge a bunch of Recurrent Neural Network (RNN) methods such as LSTM and GRU capturing the temporal variation pattern
\cite{ma2015long, tian2015predicting, dcrnn}. Also, 1-D Convolution \cite{guo2019attention} and its enhancement with Dilated Causal Convolution \cite{yu2017spatio, yu2015multi} have also achieved good performance with outstanding efficiency.  
\emph{For spatial pattern capture}, 
DeepST \cite{zhang2016dnn} and ST-ResNet \cite{strn} are representative efforts made to enhance RNN with CNN to respectively model the temporal and spatial correlation. 
Besides, 
STGCN \cite{yu2017spatio} and DCRNN \cite{dcrnn} firstly propose to describe spatial relationship in  
\zzj{\STDM} with graph structure. 
Our \baby framework hires a spatio-temporal operation set with advanced and efficient spatio-temporal operations, and assigns operation automatically, which considers spatial and temporal dependency comprehensively.

\subsection{Spatio-temporal Multi-Task Learning}
Spatio-temporal multi-task learning methods could be referred to two main lines of researches, \ie spatio-temporal multi-task learning and multi-variate time series prediction. \textit{For spatio-temporal multi-task learning,}
MDL \cite{zhang2020taxi} is one of the earliest endeavours, which incorporates convolutional neural network to solve traffic node flow and edge flow jointly. Zhang \etal propose a full LSTM method to predict traffic in and out flow together \cite{zhang2019flow}. 
{Other STMTL works includes MasterGNN \cite{han2021joint}, MT-ASTN \cite{wang2020multi}, GEML \cite{wang2019origin}, \etc}
\zzj{These models are restricted to solving two specific tasks and suffer from poor generality.}
\textit{For multi-variate time series prediction, }
MTGNN hires graph neural network for multivariate time series data prediction \cite{wu2020connecting}. DMVST-Net \cite{yao2018deep} consider multivariate time series in temporal, spatial and semantic views. 
GTS \cite{shang2021discrete} incorporates a probabilistic graph model and achieves an efficient approach for graph structure learning.
\zzj{This line of models demand human efforts for new settings due to the highly-specified architecture. Our AutoSTL is the first attempt to handle flexibly multiple spatio-temporal tasks. Its multi-task framework and shared module well exploit the attributes’ relationship. Besides, it assigns modules and hyperparameters automatically for different settings and achieves good generality.
}


\section{Conclusion}
In this paper, we present a self-adaptive framework to model multiple spatio-temporal tasks effectively.
We present a spatio-temporal operation set as candidate operation.
A scalable architecture consisting of extendable \babylayers is proposed, where each layer is composed of task-specific and shared modules.
To further enhance the multi-task learning, we employ a fusion mechanism to fuse multiple task features.
In order to support flexible combinations of multiple tasks and data, we assign operations in module and fusion weight by AutoML.
Our proposed method is the first to solve spatio-temporal multi-task learning automatically.

\zzj{In terms of the physical application of spatio-temporal prediction in today's life, our method could be easily extended to other domains such as weather and environment, public safety, human mobility, \etc In the future, we will continue discovering its potential efficacy on more applications.}


\section{Acknowledgments}
This research was partially supported by APRC - CityU New Research Initiatives (No.9610565, Start-up Grant for New Faculty of City University of Hong Kong), SIRG - CityU Strategic Interdisciplinary Research Grant (No.7020046, No.7020074), HKIDS Early Career Research Grant (No.9360163), Huawei Innovation Research Program, Ant Group (CCF-Ant Research Fund), and 
\zzj{the Fundamental Research Funds for the Central Universities, JLU}. 
Junbo Zhang is supported by the National Natural Science Foundation of China (62172034) and the Beijing Nova Program (Z201100006820053). 
Hongwei Zhao is funded by the Provincial Science and Technology Innovation Special Fund Project of Jilin Province, grant number 20190302026GX, Natural Science Foundation of Jilin Province, grant number 20200201037JC, the Fundamental Research Funds for the Central Universities for JLU.

\bibliography{7aaai23}

\end{document}